\documentclass[10pt,conference]{IEEEtran}
\IEEEoverridecommandlockouts

\usepackage{cite}
\usepackage{amsmath,amssymb,amsfonts}
\usepackage{graphicx}
\usepackage{textcomp}
\usepackage{xcolor}
\def\BibTeX{{\rm B\kern-.05em{\sc i\kern-.025em b}\kern-.08em
    T\kern-.1667em\lower.7ex\hbox{E}\kern-.125emX}}

\usepackage{booktabs}
\usepackage{todonotes}
\usepackage[acronym]{glossaries}
\usepackage{orcidlink}
\usepackage[capitalise]{cleveref}
\usepackage{url}

\usepackage{xcolor}
\usepackage{soul}

\ifCLASSOPTIONcompsoc
\usepackage[caption=false,font=normalsize,labelfont=sf,textfont=sf]{subfig}
\else
\usepackage[caption=false,font=footnotesize]{subfig}
\fi

\usepackage[ruled,norelsize]{algorithm2e}
\makeatletter
\newcommand{\removelatexerror}{\let\@latex@error\@gobble}
\makeatother

\usepackage{pifont} 
\newcommand{\cmark}{\checkmark}

\newacronym{moo}{MOO}{Multi-objective optimization}
\newacronym{qubo}{QUBO}{Quandratic Unconstrained Binary Optimization}
\newacronym{mqubo}{mQUBO}{Multi-Objective Quadratic Unconstrained Binary Optimization}
\newtheorem{theorem}{Theorem}

\DeclareMathOperator*{\argmin}{arg\,min}

\newcommand{\mP}{\mathbb{P}}
\newcommand{\mE}{\mathbb{E}}

\newcommand{\sols}{\boldsymbol{S}}
\newcommand{\solsets}{\mathcal{S}}
\newcommand{\graph}{\boldsymbol{G}}

\newcommand{\bw}{\boldsymbol{w}}

\newcommand{\utopia}{\boldsymbol{z}^{\text{ideal}}}
\newcommand{\nadir}{\boldsymbol{z}^{\text{nadir}}}
\newcommand{\refp}{\boldsymbol{z}^{\text{ref}}}
\newcommand{\desire}{\boldsymbol{z}^{\text{desire}}}

\newcommand{\bQ}{\boldsymbol{Q}}
\newcommand{\bF}{\boldsymbol{F}}
\newcommand{\bx}{\boldsymbol{x}}

\newcommand{\by}{\boldsymbol{y}}
\newcommand{\bX}{\boldsymbol{X}}

\newcommand{\cX}{\mathcal{X}}
\DeclareMathOperator{\var}{Var}
\DeclareMathOperator{\cov}{Cov}
\DeclareMathOperator{\betaD}{Beta}
\newcommand{\sd}{\sigma}

\newcommand\copyrighttext{%
  \footnotesize \textcopyright \the\year{} IEEE. Personal use of this material is permitted. Permission from IEEE must be obtained for all other uses, including reprinting/republishing this material for advertising or promotional purposes, collecting new collected works for resale or redistribution to servers or lists, or reuse of any copyrighted component of this work in other works.}

\newcommand\copyrightnotice{%
\begin{tikzpicture}[remember picture,overlay]
\node[anchor=south,yshift=10pt] at (current page.south) {\fbox{\parbox{\dimexpr0.75\textwidth-\fboxsep-\fboxrule\relax}{\copyrighttext}}};
\end{tikzpicture}%
}

\begin{document}

\title{Standardization of Multi-Objective QUBOs}
\author{
  \IEEEauthorblockN{Loong Kuan Lee
    $^{\orcidlink{0000-0002-9967-1319}}$
  }
\IEEEauthorblockA{\textit{Fraunhofer IAIS} \\
Sankt Augustin, Germany \\
loong.kuan.lee@iais.fraunhofer.de}
\and
  \IEEEauthorblockN{Thore Gerlach}
  \IEEEauthorblockA{%
    \textit{University of Bonn}\\
    Bonn, Germany \\
    tgerlac1@uni-bonn.de}
\and
\IEEEauthorblockN{Nico Piatkowski}
\IEEEauthorblockA{\textit{Fraunhofer IAIS} \\
Sankt Augustin, Germany \\
nico.piatkowski@iais.fraunhofer.de}
}
\IEEEpubid{\makebox[\columnwidth]{
    10.1109/QCE65121.2025.00017~\copyright2025 IEEE \hfill}
  \hspace{\columnsep}\makebox[\columnwidth]{ }}
\maketitle
\IEEEpubidadjcol

\newif\ifpreprint
\preprintfalse
\ifpreprint
\copyrightnotice
\fi
\begin{abstract}
  Multi-objective optimization involving Quadratic Unconstrained Binary
  Optimization (QUBO) problems arises in various domains. A fundamental
  challenge in this context is the effective balancing of multiple
  objectives, each potentially operating on very different scales. This
  imbalance introduces complications such as the selection of
  appropriate weights to balance the different objectives. In this
  paper, we propose a novel technique for scaling QUBO objectives that
  uses an exact computation of the variance of each individual QUBO
  objective. By scaling each objective to have unit variance, we align
  all objectives onto a common scale. This allows for more balanced
  solutions to be found when combining these objectives directly, as
  well as potentially assisting in the search or choice of weights
  during scalarization. Finally, we demonstrate its advantages through
  empirical evaluations on various multi-objective optimization
  problems. Our results are noteworthy since manually selecting
  scalarization weights is cumbersome; and reliable, efficient solutions
  are scarce. Code for the paper can be found at
  \url{https://gitlab.com/lklee/qubo-standardization}.
\end{abstract}
\glsunset{qubo}
\section{Introduction}
A natural formulation for problems solvable by adiabatic quantum
computing~\cite{lucas2014,date2019} and quantum approximate optimization
algorithms~\cite{blekos2024} are \emph{\acrlong{qubo}} (\gls{qubo})
problems~\cite{punnen2022quadratic}, expressed by:
\begin{equation}
  \label{eq:qubo} \min_{\bx\in\cX} \bx^{\top}\bQ\bx
  = \min_{\bx\in\cX} \: \sum_{i=1}^{n}\bQ_{i,i}\bx_{i} +
  \sum_{i,j}\bQ_{i,j} \bx_{i}\bx_{j},
\end{equation}
where $\cX = \{0,1\}^{n}$, $\bQ$ is a symmetric real-valued $n \times n$
matrix, and $n$ denotes the number of binary variables; typically
corresponding to qubits in quantum systems. 
They are also equivalent to transverse-field Ising model Hamiltonians,
up to the change of variables $\bx_{i} = (1-\sigma^{z}_{i}) / 2$, where
$\sigma^{z}$ is the Pauli Z operator~\cite{satoshi2008}. The ground
state of the resulting Hamiltonian is then equivalent to the vector
$\bx$ that minimizes \Cref{eq:qubo}.

Despite its 
simple quadratic form, the \gls{qubo} problem is
NP-hard~\cite{pardalos1992} and serves as a unifying framework for a
broad class of combinatorial optimization tasks. Notable application
domains include chip design~\cite{gerlach2024fpga2},
finance~\cite{aguilera2024,sakuler2025,lu2024}, flight gate
assignment~\cite{stollenwerk2019flight}, and various machine learning
tasks~\cite{bauckhage2019, mucke2023feature}.


\begin{figure*}
  \centering
  \subfloat[The difference in scale of the Pareto frontier for
  $\min_{\bx} f(\bx)+g(\bx)$ before and after standardization. Pareto
  frontier was found by a brute force algorithm.]{
    \includegraphics[width=0.25\textwidth]{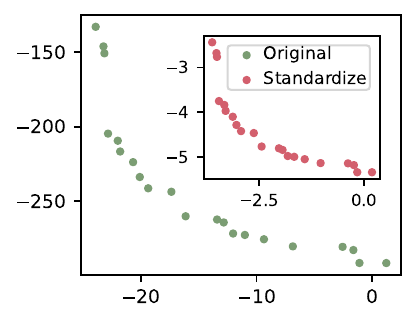}
    \label{subfig:nested-pfs}
  }
  \hfill
  \subfloat[The distribution of objective values of
  $f(\bx)=\bx^\top\bQ_{1}\bx$ (with $n=20$) where $\bx$ are sampled
  uniformly. Also indicated are the exact range of $f(\bx)$ obtained
  from solving it, the estimated range from its roof dual bound, as well
  as its exact mean and variance.]{
    \includegraphics[width=0.4\textwidth]{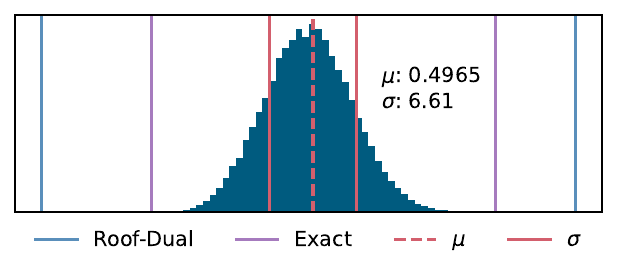}
    \label{subfig:energy-range}
  }
  \hfill
  \subfloat[The solution obtained from solving
  $\min_{\bx} f(\bx)+g(\bx)$ exactly before and after
  standardization. The histograms on the x and y axes represents the
  distribution of $f(\bx)$ and $g(\bx)$ respectively.]{
    \includegraphics[width=0.25\textwidth]{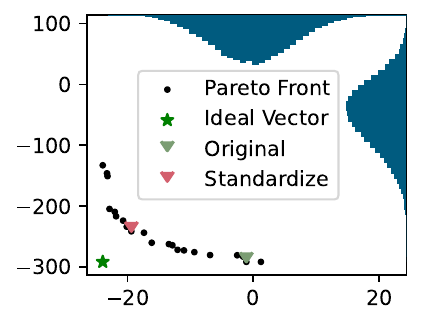}
    \label{subfig:example-sols}
  }
  \caption{Plots over the distribution (b) and the objective space (a,c)
    of the \gls{qubo} objective functions $f(\bx)=\bx^{\top}Q_{1}\bx$
    and $g(\bx)=\bx^{\top}Q_{2}\bx$. The $20\times20$ symmetric
    real-valued matrices $Q_{1}$ and $Q_{2}$ are randomly generated by
    the \texttt{qubolite} package~\cite{mucke2025} with
    \texttt{random\_state} seeds of $0$ and $1$
    respectively. Additionally $Q_{2}$ is scaled up by a factor of $10$
    as well.}
  \label{fig:intro}
\end{figure*}

\subsection{Motivation}

\gls{moo} involves optimizing two or more conflicting objectives
simultaneously~\cite{branke2008,miettinen1999nonlinear}.
This approach is essential in real-world decision-making, where
trade-offs between competing goals must be carefully balanced.  Unlike
single-objective problems, \gls{moo} yields a set of Pareto-optimal
solutions, offering decision-makers flexibility in selecting the most
suitable outcome.  Its importance is evident across diverse fields,
including vehicle routing~\cite{xu2022},
telecommunications~\cite{bouchmal2025, lin2025}, engineering
design~\cite{gunantara2018review,matsumori2022}, environmental
plannig~\cite{marcelino2021efficient}, and
finance~\cite{dachert2022,aguilera2024}.

Solution methods in MOO can be broadly categorized based on how
decision-maker preferences are incorporated: no-preference, a priori,
and a posteriori approaches~\cite{miettinen1999nonlinear,miettinen2008}.
No-preference methods aim to find representative solutions without
incorporating any decision-maker preferences, making them suitable when
such information is unavailable or deliberately excluded.  A priori
methods require the decision-maker to express their preferences in
advance, often through scalarization techniques that reduce the problem
to a single-objective formulation~\cite{gass1955,zadeh1963}.  While
effective, these methods assume that preference information is available
and well-understood beforehand, which may not always be the case.  In
contrast, a posteriori methods first approximate or compute a diverse
set of Pareto-optimal solutions, allowing the decision-maker to explore
trade-offs after the
optimization~\cite{ayodele2023,ayodele2023a,lin2024}.  Although this
reduces the need for precise prior knowledge, generating the Pareto
front can be computationally intensive.

In this work, we adopt a no-preference approach. This choice allows us
to build a simple and general foundation for demonstrating our method,
without relying on strong assumptions about decision-maker preferences.
Importantly, our approach can be seamlessly integrated with a priori or
a posteriori techniques, which can be applied subsequently to refine the
solution process based on user-defined goals or exploratory analysis.

\subsection{Problem Setup}\label{sec:intro-problem}
Recall that in \gls{moo}, we have a set of objective functions
$f_{1}(\bx),f_{2}(\bx),\ldots,f_{m}(\bx)$ we wish to optimize over
simultaneously. However, in this paper, we will focus on \gls{moo}
problems constructed from multiple \gls{qubo} problems---referred to as
\gls{mqubo} problems. Therefore, our objective functions specifically
take the form
\begin{equation}
  \label{eq:qubo-objective-function}
  \forall i\in\{1,\ldots,m\} : f_{i}(\bx) = \bx^{\top}\bQ^{(i)}\bx,
\end{equation}
where $\bx$ is a binary vector of length $n$ and $\bQ$ is a symmetric real-valued $n\times n$ matrix, similar to \Cref{eq:qubo}. 

One approach to finding a representative solution in a no-preference
setting is the method of global criterion~\cite{miettinen2008}. The
global criterion involves minimizing the distance between the objective
vector of a solution and some desirable point in the objective space.  A
natural choice of this desirable point is the ideal vector
$\utopia$ and is constructed from the minimum value of each objective function~\cite{yu1973},
\begin{gather}
  \label{eq:global-criterion}
  \min_{\bx\in\cX} \left\|f_{i}(\bx) - \utopia_{i}\right\|_{p},
\end{gather}
where $\utopia=[\min_{\bx} f_{1}(\bx),\ldots,\min_{\bx} f_{m}(\bx)]$.
By using the $L_{1}$-metric ($p=1$) in the global criterion, we can transform
\Cref{eq:global-criterion} to a linear scalarization of our objective
functions,
\begin{equation}
  \label{eq:global-criterion-L1}
  \min_{\bx\in\cX} \sum_{i=1}^{m}\left|f_{i}(\bx) - \utopia_{i}\right|
  =\min_{\bx\in\cX}\left( \sum_{i=1}^{m}f_{i}(\bx)\right)-c,
\end{equation}
with $c=\sum_{i=1}^{m}\utopia_{i}$, since the terms in the absolute values are always positive by the
definition of $\utopia$.  Therefore, this approach to finding a solution
with no-preference is directly applicable to \gls{mqubo} as we can now
just minimize the following \gls{qubo} problem:
\begin{equation}
  \min_{\bx\in\cX} \sum_{i=1}^{m}f_{i}(\bx)
  = \min_{\bx\in\cX} \bx^{\top} \Biggl(\sum_{i=1}^{m}\bQ^{(i)}\Biggr) \bx
  = \min_{\bx\in\cX} \bx^{\top} \tilde{\bQ} \bx
  \label{eq:mqubo}
\end{equation}
to find a Pareto optimal solution.

However, in general, scalarization methods for \gls{moo} can be hampered
by objective functions with differing scales as this may lead to some
subset of our objective functions being over-favored during
optimization. A common approach to ensure that that the objective
functions are in similar scales is to normalize each objective function
by dividing them with their respective
ranges~\cite{deb2002,he2021,miettinen1999nonlinear,miettinen2008},
\begin{equation}
  \label{eq:2}
  \tilde{f}_{i}(\bx) = \frac{f_{i}(\bx) - \utopia_{i}}{\nadir_{i} - \utopia_{i}},
\end{equation}
where $\nadir=[\max_{\bx} f_{1}(\bx),\ldots,\max_{\bx} f_{m}(\bx)]$.
However getting the exact range of a general \gls{qubo} objective
function is equivalent to solving it exactly, and therefore impractical in the
case of \gls{mqubo} problems. It is possible to estimate a lower and
upper bound on the objective value of a \gls{qubo} problem by using relaxation techniques, e.g., by calculating the \emph{roof dual bound}\footnote{See \Cref{sec:exp} for a short
  description of the roof dual bound.} of the
problem~\cite{boros2008max} or by using \emph{semi-definite programming} \cite{goemans1995improved}. However, as we can see in
\Cref{subfig:energy-range}, these bounds might not be tight, and
therefore might end up causing the objective function to be
under-weighted instead.

\subsection{Contribution}
Therefore, we propose the use of standardization as a means to ensure that the objective functions in a \acrfull{mqubo} are within the
same scale instead; the goal of which is to encourage the solution of
the scalarization in \Cref{eq:mqubo} to be closer to the ideal point
$\utopia$. See \Cref{subfig:example-sols} for an example.

Our main contributions are
\begin{itemize}
\item a closed-form expression of the mean and variance for a \gls{qubo}
  problem's objective values,
\item an algorithm to compute said variance with a computational
  complexity of $\mathcal{O}(n^{3})$ where $n$ is the number of binary variables
  $|\bx|$,
\item and finally experiments showing the ability for standardization to
  result in more balanced solutions in the no-preference setting
  compared to using the original un-scaled objectives, or
  with normalization using roof dual bound estimated ranges.
\end{itemize}

The details of this approach, the algorithm, and their derivation can be
found in \Cref{sec:method}. Then in \Cref{sec:exp}, we compare the
no-preference solution found by our approach with the solutions found
when using the roof dual bound for normalizing \gls{qubo} objective
functions, as well when no scaling approaches are used. We finally
conclude this paper in \Cref{sec:conc} with some closing thoughts and
ideas for future work.

\section{Standardizing QUBO Objective Functions}\label{sec:method}
Standardization is a commonly used method in statistics to re-scale a
random variable such that it has a mean of $0$ and a standard deviation
of $1$.
To apply it to our problem, first assume that the solution vector $\bx$
can be modeled as being sampled from some probability distribution
$\mP_{\bX} : \cX \to \mathbb{R}_{[0,1]}$, where $\bX$ is
the \emph{random vector} of binary random variables. We can then
standardize the objective functions in our \gls{mqubo} problem with
respect to $\mP_{\bX}$,
\begin{equation}
  \label{eq:standardize-mqubo}
  \begin{aligned}
    &\argmin_{\bx\in\cX}\sum_{i=1}^{m}
    \frac{f_{i}(\bx) - \mathbb{E}_{\bX}[f_{i}(\bX)]}
    {\sd_{\bX}(f_{i}(\bX))}\\
    &=\argmin_{\bx\in\cX}\Biggl(\sum_{i=1}^{m} \frac{f_{i}(\bx)}
    {\sd_{\bX}(f_{i}(\bX))}\Biggr) - c.
  \end{aligned}
\end{equation}
where $c=\sum_{i=1}^{m}\mathbb{E}_{\bX}[f_{i}(\bX)]$ is a
constant.

Therefore, the standardization approach only requires
dividing the \gls{qubo} matrices by the standard deviation $\sd_{\bX}(f_{i}(\bX))$ of $f_{i}(\bx)=\bx^{\top} \bQ^{(i)}\bx$ with respect to some distribution
$\mP_{\bX}$.
Two questions arise: what is $\mP_{\bX}$
and how is $\sd_{\bX}(f_{i}(\bX))$ computed?

Which distribution $\mP_{\bX}$ is correct for a specific scenario 
depends on various unknown factors, e.g., the actual \gls{mqubo} 
matrices, constraints, the underlying combinatorial problems, and 
so on. In the presence of this high degree of uncertainty, we choose 
to maximize the entropy of the underlying distribution. That is, we
assume $\mP_{\bX}$ is a uniform distribution over $\cX=\{0,1\}^n$. 

There are two interpretations on what a uniform $\mP_{\bX}$ means. The
first is that each candidate solution in $\cX$ has a probability of
$2^{-n}$ to be sampled. The second interpretation is that all random
variables in $\bX$ are independent Bernoulli variables distributed with
a success probability of $p=0.5$. To start off this section, we will use
the first interpretation to derive a closed-from expression of the mean
and variance for our objective functions.


\begin{theorem}[Mean and variance of \gls{qubo} with uniform $\bX$]
  \label{thm:uniform-mean-var}
  Assume we have the objective function
    $f(\bx) = \bx^{\top} \bQ \bx$ 
  where $\bx$ is a binary vector of length $n$ and $\bQ$ is an
  $n\times n$ symmetric real-valued matrix. Furthermore, let $\mP_{\bX}$
  be the uniform distribution over the space of all possible $n$-length
  binary vectors $\cX$,
    $\forall \bx \in \cX : \mP_{\bX}(\bx) = 2^{-n}$.
  Then the first two moments of the objective function can be computed in closed-form in time $\mathcal{O}(n^{4})$ via 
  \begin{equation}
    \label{eq:uniform-mean}
    \mE[f(\bX)]
    =\frac{1}{2} \sum_{i=1}^n\bQ_{i,i} + \frac{1}{4}
    \sum_{i=1}^n\sum_{j\not=i}^n\bQ_{i,j}
  \end{equation}
and
  \begin{equation}
    \label{eq:uniform-moment-2}
    \mE[f(\bX)^2] =  \sum_{i,j,k,l=1}^{n}
    \bQ_{i,j}\bQ_{k,l}2^{-|\{i,j,k,l\}|}.
  \end{equation}
\end{theorem}
\begin{IEEEproof}
  Throughout this proof, any sum over the indices $i,j,k,l$ are implied
  to be from $1$ to $n$ unless indicated otherwise. We shall now start
  by deriving the mean of $f(\bX)$:
  \begin{align*}
    \mE[f(\bX)] 
    &= \sum_{\bx\in\cX} \mP(\bx) \; \bx^\top \bQ \bx
    = \sum_{\bx\in\cX} \mP(\bx) \sum_{i=1}^{n}\sum_{j=1}^{n}
      \bQ_{i,j} \bx_{i}\bx_{j}\\
    &=  2^{-n} \sum_{i=1}^n\sum_{j=1}^n\bQ_{i,j}
      \sum_{\bx\in\cX} \bx_i\bx_j\\
    &=  2^{-n} \sum_{i=1}^n\sum_{j=1}^n\bQ_{i,j} \times
      \begin{cases}
        2^{n-1}&\text{ if }i=j\\
        2^{n-2}&\text{otherwise}
      \end{cases}\\
    &=\sum_{i=1}^n\sum_{j=1}^n\frac{\bQ_{i,j}}{2^{|\{i,j\}|}} 
    =\frac{1}{2} \sum_{i=1}^n\bQ_{i,i} + \frac{1}{4}
      \sum_{i=1}^n\sum_{j\not=i}^n\bQ_{i,j}
  \end{align*}
  where the result of the sum $\sum_{\bx\in\cX} \bx_i\bx_j$ depends on
  the number of dimensions---i.e. variables---in $\bX$ that are fixed.
  We can then use this trick in a similar way to compute the second
  moment of $f(\bX)$ and therefore the variance,
  \begin{align*}
  \mE[f(\bX)^2]
  &= \sum_{\bx} \mP(\bx) \left( \sum_{i=1}^n\sum_{j=1}^n\bQ_{i,j} \bx_i\bx_j \right)^2\\
  &= \sum_{\bx} \mP(\bx) \sum_{i,j,k,l=1}^{n} \bQ_{i,j}\bQ_{k,l} \bx_i\bx_j \bx_k\bx_l\\
  &=  \sum_{i,j,k,l=1}^{n}\bQ_{i,j}\bQ_{k,l}2^{-|\{i,j,k,l\}|},
  \end{align*}
  thus proving the equality in \Cref{eq:uniform-moment-2}.
\end{IEEEproof}

While the variance derivation in \Cref{thm:uniform-mean-var} is
succinct, direct computation of the sum in \Cref{eq:uniform-moment-2}
has a time complexity of $\mathcal{O}(n^{4})$, which becomes
computationally expensive for large $n$. To address this,
\Cref{thm:variance} derives a closed-form expression for the variance of
$f(\bX)$ with a reduced computational complexity of
$\mathcal{O}(n^{3})$, which can lead to significant speedups for large
$n$.

\begin{theorem}[Variance of \gls{qubo} with independent Bernoulli
  distributions $\mathcal{B}(0.5)$]
  \label{thm:variance}
  Assume we have the objective function
    $f(\bx) = \bx^{\top} \bQ \bx$
    where $\bx$ is a binary vector of length $n$ and $\bQ$ is a
    symmetric real-valued matrix. Then further assume that $\bX$ is a
    vector of independent Bernoulli random variables with success
    probability $p=0.5$,
    $\forall i \in \{1,2,\ldots,n\} : \bX_{i} \sim \mathcal{B}(0.5)$.
  The variance of $f(\bX)$ is then:
  \begin{equation}
    \label{eq:bern-var}
    \begin{aligned}
     &\var(f(\bX))\\
      &=\sum_{i} \frac{1}{4}\bQ_{i,i}^{2} + \frac{1}{8}\Biggl(
        \sum_{k\neq i} \bQ_{i,i} \cdot [\bQ_{k,i} + \bQ_{i,k}]
        \Biggr)\\
      &\quad+\sum_{i}\sum_{j\neq i} \bQ_{i,j}\Biggl(
        \frac{3}{16} \Bigl(\bQ_{i,j} + \bQ_{j,i}\Bigr)
        + \frac{1}{8} \Bigl(\bQ_{i,i} + \bQ_{j,j}\Bigr)
        \Biggr)\\
      &\quad+\sum_{i}\sum_{j\neq i} \bQ_{i,j}
        \sum_{k\neq\{i,j\}}\frac{1}{16}
        \Bigl(\bQ_{k,i} + \bQ_{k,j} + \bQ_{i,k} + \bQ_{j,k}\Bigr).
    \end{aligned}
  \end{equation}
\end{theorem}
\begin{IEEEproof}
  First notice that the product of any Bernoulli variable with itself
  results in the same random variable,
  $\bX_{i}\cdot\bX_{i}=\bX_{i}$. Furthermore, the product of two
  idependent Bernoulli random variables
  $\bX_{i},\bX_{j}\sim\mathcal{B}(p)$ results in the new Bernoulli
  random variable $\bX_{i}\cdot\bX_{j}\sim\mathcal{B}(p^{2})$. Then recall
  the following equivalence between the variance of a sum of random
  variables and sum of their pairwise covariance, also known as
  Bienaym\'{e}'s identity,
  \begin{align*}
    \var(f(\bX))
    =&\var\biggl(\sum_{i,j}\bX_{i}\bX_{j}\bQ_{i,j}\biggr)\\
    =& \sum_{i,j}\sum_{k,l}\bQ_{i,j}\bQ_{k,l}
       \cov(\bX_{i}\bX_{j},\bX_{k}\bX_{l}).
  \end{align*}
  Note that for $\cov(\bX_{i}\bX_{j},\bX_{k}\bX_{l})$ to be non-zero,
  $\bX_{i}\bX_{j}$ and $\bX_{k}\bX_{l}$ needs to share a common variable
  since if they do not,
  \begin{align*}
    &\cov(\bX_{i}\bX_{j},\bX_{k}\bX_{j})\\
    &=\mE[\bX_{i}\bX_{j}\bX_{k}\bX_{j}] -
    \mE[\bX_{i}\bX_{j}]\mE[\bX_{k}\bX_{j}]\\
    &=2^{-|\{i,j\}|-|\{k,l\}|} - 2^{-|\{i,j\}|}\cdot2^{-|\{k,l\}|}
    =0.
  \end{align*}
  Therefore, one approach we can use to compute the variance is to
  enumerate all the cases when $\cov(\bx_{i}\bx_{j},\bx_{k}\bx_{l})$ is
  non-zero.
  \begin{enumerate}
  \item When $i=j$, and
    \begin{enumerate}
    \item $k=l=i$, the covariance is $1/4$.
    \item $k=i$ but $l\neq i$, the covariance is $1/8$.
    \item $l=i$ but $k\neq i$, the covariance is $1/8$.
    \end{enumerate}
  \item When $i\neq j$, and
    \begin{enumerate}
    \item $k=i$ and $l=j$, the covariance is $3/16$.
    \item $k=j$ and $l=i$, the covariance is $3/16$.
    \item $k=l=i$ or $k=l=j$, the covariance is $1/8$.
    \end{enumerate}
  \item When $i\neq j$, and
    \begin{enumerate}
    \item $k=i$ and $l\notin \{i,j\}$, the covariance is $1/16$.
    \item $k=j$ and $l\notin \{i,j\}$, the covariance is $1/16$.
    \item $l=i$ and $k\notin \{i,j\}$, the covariance is $1/16$.
    \item $l=j$ and $k\notin \{i,j\}$, the covariance is $1/16$.
    \end{enumerate}
  \end{enumerate}
  As a result, we can now compute the variance of $f(\bX)$ by just
  taking the sum over the indices where we know the covariance term is
  non-zero and grouping terms with the same covariance together,
  resulting in \Cref{eq:bern-var}.
\end{IEEEproof}

Furthermore, when implementing the expression in \Cref{eq:bern-var}, it
is also possible to further simplify the terms in the sum. This results
in a concise algorithm for the computation of the variance for
any \gls{qubo} objective function---as seen in \Cref{alg:var}---with
a computational complexity of $\mathcal{O}(n^{3})$.

\begin{figure}
  \removelatexerror
  \begin{algorithm}[H]
    \caption{Fast QUBO Variance in $\mathcal{O}(n^{3})$}
    \KwIn{\gls{qubo} matrix $Q$}
    \KwOut{Variance $v$}
    $n \leftarrow$ number of rows in $Q$\;
    $v \leftarrow 0$\;
    \For{$i \leftarrow 1$ \KwTo $n$}{
      \For{$j \leftarrow 1$ \KwTo $n$}{
        \If{$i \neq j$}{
          $v \leftarrow v + \bQ_{i,j} (\bQ_{i,j} + \bQ_{j,i}) / 16$\;
        }
        \For{$k \leftarrow 1$ \KwTo $n$}{
          $v \leftarrow v + \bQ_{i,j} (\bQ_{k,i} + \bQ_{k,j} +\bQ_{i,k} +\bQ_{j,k}) / 16$\;
        }
      }
    }
    \Return $v$\;
  \end{algorithm}
  \caption{Algorithm for computing the variance of a \gls{qubo}
    objective function.}
  \label{alg:var}
\end{figure}

\section{Experiment}\label{sec:exp}
We now wish to showcase the applicability of our approach of using
standardization for scaling the objective functions in \gls{mqubo}
problems. Recall from \Cref{sec:intro-problem} we argued that the main
benefit of scaling the objective functions is to discourage any single
objective function from being over-prioritized in the final scalarized
compound objective function. Therefore in this section, we will compare
the solutions we get from different scaling approaches when solving the
equal weighted linear scalarization of our objective
functions. Specifically we will compare,
\begin{enumerate}
\item a scalarization where none of the objective functions are scaled,
\item an approach that normalizes the objectives by estimating
  the range of each \gls{qubo} objective function via their roof dual
  bound, and
\item our approach of scaling via standardization.
\end{enumerate}

The \emph{roof dual bound}~\cite{boros2008max} is a relaxation technique
for QUBO problems that provides a tight lower bound by solving a related
continuous optimization problem.  It leverages roof duality theory to
identify variables that can be fixed to binary values---known as
persistent variables---based on the structure of the quadratic
objective.  This is often achieved through a max-flow formulation.
Variables with strong persistency are guaranteed to have the same values
in any optimal solution, enabling effective preprocessing and problem
size reduction with a worst-case runtime of $\mathcal O(n^3)$. In our
experiments, we specifically use the implementation found in
qubolite~\cite{mucke2025}.

\subsection{QUBO Problems}
We will now lay out the \gls{mqubo} problems and methodology we will use
to compare the discussed scaling methods. To start we first define 4
different \gls{qubo} problems and their respective objective functions
for use in our experiments. The following max cut and subset sum
problems~\cite{karp1972} are all defined on the vertices $V(\graph)$ of
a random Barabasi-Albert graph $G$. Each \gls{qubo} problem utilizes
the connectivity of the initial graph $G$ in its construction.
\begin{enumerate}
\item MC$[0,1]$: A max cut problem where the weights on the graph
  $\graph$ are sampled from the beta distribution,
  $\forall (i,j) \in E(\graph)$:
    \begin{equation*}
        \bw_{i,j}^{(P)} \sim \betaD(0.2, 0.8)
    \end{equation*}
  \item MC$\{0,1\}$: A max cut problem defined on the complete graph
    over the vertices of $\graph$, $K_{\graph}$. The weights of each
    edge on the complete graph are sampled from a Bernoulli distribution
    with equal probability of sampling $0$ or $1$. However, the weight of
    edges that already exist in $E(\graph)$ are always $1$:
    \begin{equation*}
      \forall (i,j) \in E(K_{\graph}):
      \begin{cases}
        \bw_{i,j}^{(B)} = 1 & (i,j) \in E(\graph)\\
        \bw_{i,j}^{(B)} \sim \mathcal{B}(0.5) & \text{otherwise}\\
      \end{cases}
    \end{equation*}
  \item MC$[1,5]$: A max cut problem defined on the complete graph over
    the vertices of $\graph$, $K_{\graph}$. The weights of each edge on
    the complete graph are sampled from a discrete uniform distribution
    from $1$ to $5$, $\mathcal{U}\{1,5\}$. However, the weight of edges
    that already exist in $E(\graph)$ are always $5$:
   \begin{equation*}
    \forall (i,j) \in E(K_{\graph}):
      \begin{cases}
        \bw_{i,j}^{(Z)} = 5 & (i,j) \in E(\graph)\\
        \bw_{i,j}^{(Z)} \sim \mathcal{U}\{1,5\} & \text{otherwise}\\
      \end{cases}
    \end{equation*}
  \item SubSum: A subset sum problem defined on the vertices of
    $\graph$, $V(\graph)$, where the weights of each vertex is the
    number of neighbors it has,
    \begin{equation*}
      \forall v \in V(\graph) :
      \bw_{v}^{(S)} = |N_{\graph}(v)|.
    \end{equation*}
    We also use a quarter of the total vertex weights as the target
    sum for the problem,
    \begin{equation*}
    \tau = \frac{1}{4}\sum_{v\in V(\graph)}\bw_{v}^{(S)}.
    \end{equation*}
  \end{enumerate}
The max-cut problems MC$[0,1]$, MC$\{0,1\}$, and MC$[1,5]$ have
objective functions with the following form~\cite{boros1991}:
\begin{equation*}
  f(\bx) = \sum_{i=1}^{n}\sum_{j=i+1}^{n} 2\bw_{i,j} \bx_{i}\bx_{j}
  - \sum_{i=1}^{n}\bx_{i}\Biggl(\sum_{j=1,j\neq i}^{n}\bw_{i,j}\Biggr)
\end{equation*}
where $\bw \in \{\bw^{(P)}, \bw^{(B)}, \bw^{(Z)}\}$. On the other hand,
the subset sum problem has the following objective
function~\cite{biesner2022}:
\begin{equation*}
    f_{S}(\bx)=\sum_{ij}\bw_{i}^{(S)}\bw_{j}^{(S)}\bx_{i}\bx_{j}-
    2\tau\sum_{i} \bw_{i}^{(S)}\bx_{i}.
\end{equation*}

These \gls{qubo} objectives can be motivated by the real-life problem of
marketing a multiplayer video game and selecting a subset of influencers
to be part of the marketing campaign. Here MC$[0,1]$ represents the
friendship between influencers and how likely they are to get another
influencer to play the game if they are part of the
campaign. MC$\{0,1\}$ and MC$[1,5]$ on the other hand indicates if any
two influencer share a common language and how compatible their
timezones are respectively. And finally SubSum represents the budget we
want to use up in the campaign.


\subsection{Experimental setup}
For our experiments we instantiate instances of our \gls{qubo}
problems with a random Barabasi-Albert graph with $1000$ nodes and a
connectivity of $2$ using the NetworkX
library~\cite{SciPyProceedings_11} with a seed of $1$. The distributions
used for problem generation are initialized using
numpy~\cite{harris2020array} with seeds of $1$ as well.

We then choose all possible combinations of the four \gls{qubo} problems
leading to $\binom{4}{2} + \binom{4}{3} + \binom{4}{4} = 11$ different
combinations, and therefore \gls{mqubo} problems for our
experiments. For each \gls{mqubo} problem, we use the 3 different
approaches for objective function scaling to create 3 different
scalarizations of the \gls{mqubo} problem, resulting in a total of $33$
different scalarized \gls{qubo} problems to solve. We also define the
operator $\bF$ for each of the $11$ \gls{mqubo} problems such that
$\bF(\bx)$ returns an $m$-length vector in objective space, where $m$ is
the number of objectives.

These scalarizations are then optimized using the digital annealer from
Fraunhofer IAIS which implemets an evolutionary algorithm for solving
\gls{qubo} problems on an FPGA~\cite{MuckePM19}. The digital annealer is
given a time limit of $2$ seconds per run to solve each given scalarized
problem. We then run the digital annealer $20$ times for each scalarized
problem, before returning a set of solutions found over those $20$ runs.

We filter this set of solutions to find the set of non-dominated
solutions $\sols$ found by the digital annealer. A solution $\bx$ is
said to \textit{dominate} another solution $\by$ if every element of
$\bx$ is less than or equal to the corresponding element of $\by$, and
at least one element in $\bx$ is strictly less than the corresponding
element in $\by$~\cite{miettinen2008},
\begin{equation}
  \label{eq:dominace}
  (\forall i\in\mathbb{N}_{[1,n]}\; \bx_{i} \leq \by_{i}) \wedge
  (\exists j\in\mathbb{N}_{[1,n]}\; \bx_{j} < \by_{j}).
\end{equation}
This entire process is repeated $20$ times to get $20$ sets of non-dominated solutions,
$\solsets=\{\sols_{1},\ldots,\sols_{20}\}$.
\begin{table*}
  \centering
  \caption{Mean Hypervolume of the set of non-dominated solutions found
    by the digital annealer over 20 runs. Highest means for each
    \gls{mqubo} problem are in bold.}
  \label{tab:hypervolume}
  \begin{tabular}{cccc rr rr rr}
    \toprule
    \multicolumn{4}{c}{\gls{qubo} Problems}
    & \multicolumn{2}{c}{Original} & \multicolumn{2}{c}{Roof Dual} & \multicolumn{2}{c}{Standardization} \\
    MC[0,1] & MC\{0,1\} & MC[1,5] & SubSum & Mean & Std Dev & Mean & Std Dev & Mean & Std Dev \\
    \midrule
    \cmark & \cmark &   &   & 3.005e+06 & 2.276e-09 & 3.18e+06 & 4.787e+04 & \textbf{5.071e+06} & 1.818e+04\\
    \cmark &   & \cmark &   & 1.128e+07 & 2.484e-08 & 1.232e+07 & 3.847e+04 & \textbf{1.849e+07} & 5.265e+04\\
    \cmark &   &   & \cmark & 9.894e+08 & 1.621e+06 & 1.106e+09 & 9.095e+06 & \textbf{1.666e+09} & 6.928e+06\\
    & \cmark & \cmark &   & \textbf{3.038e+04} & 2.44e-11  & 3.029e+04 & 5.699e-11 & 3.029e+04 & 5.699e-11\\
    & \cmark &   & \cmark & 1.644e+11 & 4.971e+07 & 1.635e+11 & 0.0002572 & \textbf{1.973e+11} & 2.286e+08\\
    &   & \cmark & \cmark & 3.786e+11 & 2.973e+07 & 3.793e+11 & 0.0007165 & \textbf{4.768e+11} & 1.935e+09\\
    \cmark & \cmark & \cmark &   & 1.828e+10 & 2.17e-05  & 6.281e+09 & 6.523e+06 & \textbf{2.368e+10} & 2.322e-05\\
    \cmark & \cmark &   & \cmark & 1.18e+14  & 1.65e+11  & 3.486e+14 & 6.299e+12 & \textbf{4.567e+14} & 2.129e+12\\
    \cmark &   & \cmark & \cmark & 2.237e+14 & 7.501e+09 & 5.879e+14 & 1.423e+13 & \textbf{8.484e+14} & 4.755e+12\\
    & \cmark & \cmark & \cmark & 2.353e+16 & 50.35     & 7.004e+16 & 24.00     & \textbf{8.04e+16}  & 59.53\\
    \cmark & \cmark & \cmark & \cmark & 7.934e+18 & 1.146e+14 & 6.45e+19  & 2.484e+16 & \textbf{8.099e+19} & 7.914e+15\\
    \bottomrule
  \end{tabular}
\end{table*}

\subsection{Performance Measure: Hypervolume Indicator}
\begin{figure}
  \centering
  \includegraphics[width=0.35\textwidth]{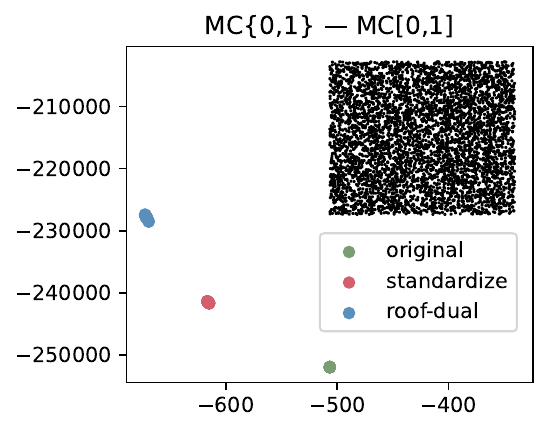}
  \caption{Example of reference points (black) used for the Hypervolume
    calculations---that are later averaged over---in 2D objective space
    over problems MC$\{0,1\}$ and MC$[0,1]$.}\label{fig:hv-ex}
\end{figure}
The measure we use to compare the solutions in $\sols\in\solsets$
between the different scaling methods is to compute the hypervolume of
$\sols$ found from optimizing the scalarizations corresponding to each
method. Hypervolume is a standard approach for evaluating solution sets
to \gls{moo} problems and essentially measures the quality and diversity
of a set of non-dominated solutions. It does so by finding the size of
the objective space dominated by the objective vector of at least one
solution $\bF(\bx), \bx\in\sols$ and bounded by a chosen \emph{reference
  point}~\cite{zitzler1998}. The \emph{reference point} used must be
dominated by all the solutions in $\sols$ in objective space. Therefore,
the larger the hypervolume indicator is, the greater the region the
solution set $\sols$ dominates in objective space; implying a solution
set of higher quality.

For our experiments, we actually use a set of $10000$ uniformly
generated \emph{reference points} over the hypercube defined by two
points, $\refp$ and $2\cdot\refp - \desire$, and average the computed
hypervolume over these \emph{reference points}. Here the point $\refp$
is unique to each of the 11 \gls{mqubo} problems and is the element-wise
maximum in objective space over all
$\sols\in(\solsets_{N}\cup\solsets_{S}\cup\solsets_{O})$, where
$(\solsets_{N},\solsets_{S},\solsets_{O})$ corresponds to the set of
sets of solutions from optimizing the scalarization with normalization,
standardization, and with no scaling respectively. On the other hand,
$\desire$ represents the element-wise minimum over the same set of
solutions instead. The hypercube defined between $\refp$ and
$2\cdot\refp - \desire$ is then just a translated copy of the hypercube
defined by the points $\desire$ and $\refp$. An example of these
randomly generated \emph{reference points} for a two objective problem
can be found in \Cref{fig:hv-ex}.

\subsection{Results}
\begin{table}
  \centering
  \caption{The roof dual range and standard deviation for each
    \gls{qubo} problem used in our experiments.}\label{tab:range}
  \begin{tabular}{l r r}
    \toprule
    QUBO & Roof Dual Range & Std. Deviation\\
    \midrule
    MC[0,1]	 & 7.578e2 & 2.235e1\\
    MC\{0,1\}& 2.520e5 & 4.605e3\\
    MC[1,5]	 & 1.507e6 & 2.753e4\\
    SubSum	 & 2.680e7 & 4.216e5\\
    \bottomrule
  \end{tabular}
\end{table}

From \Cref{tab:hypervolume}, we can see that standardization almost
always score higher on the hypervolume indicator, implying that the
solutions we get from standardizing the objective functions of a
\gls{mqubo} before equal weighted scalarization seems to be providing
more balanced trade-offs between the different objectives.

However, for the \gls{mqubo} problem with the objectives MC$\{0,1\}$ and
MC$[1,5]$, the solution found for the scalarization without any scaling
performs best. That said, the hypervolume indicators for this
\gls{mqubo} problem are very close to each other, with a maximum
difference of $0.2962$\%. We hypothesize that the reason for the similar
performances is partly due to MC$\{0,1\}$ and MC$[1,5]$ being defined
very similarly; they are both max-cut problems on the complete graph
where edges that were already in the original graph are given the
largest weight. Furthermore, from \Cref{tab:range}, we can see that the
roof dual range and standard deviation of MC$\{0,1\}$ and MC$[1,5]$ are
different by just a single order of magnitude. So we believe it is a
combination of both factors that led to the different approaches
performing similarly.

\section{Conclusion}\label{sec:conc}
When multiple objective functions have to be addressed simultaneously, 
setting approriate preferences is a notoriously hard and cumbersome process. 
No-preference methods aim to find representative solutions but implicitly 
require that all invovled objectives have a similar scale.

To this end, we proposed the use of standardization as a means to
ensure a set of \gls{qubo} objective functions are within a similar
scale---as opposed to normalization. The main benefit of using
standardization is that it can be done exactly with a computational
complexity of $\mathcal{O}(n^{3})$. We proved the computational
complexity of standardization by both deriving a closed-form expression
for the variance of a \gls{qubo} objective function when the binary
solution vector $\bx$ is sampled from a uniform distribution. We also
provided a concise algorithm for computing this variance.  Finally, we
showed that solutions obtained from equal weighted scalarization with
standardization scores higher on the hypervolume indicator on average
compared to other scaling approaches. Therefore standardization provides
a more balanced trade-off between the different objectives compared to
just using the original objective function or normalization using a
roof dual bound estimate on the range of the objectives.


\section*{Acknowledgment}
This research has been funded by the Federal Ministry of Education and
Research of Germany and the state of North-Rhine Westphalia as part of
the Lamarr-Institute for Machine Learning and Artificial Intelligence.

\newpage







\bibliographystyle{IEEEtran}

\begin{thebibliography}{10}
\providecommand{\url}[1]{#1}
\csname url@samestyle\endcsname
\providecommand{\newblock}{\relax}
\providecommand{\bibinfo}[2]{#2}
\providecommand{\BIBentrySTDinterwordspacing}{\spaceskip=0pt\relax}
\providecommand{\BIBentryALTinterwordstretchfactor}{4}
\providecommand{\BIBentryALTinterwordspacing}{\spaceskip=\fontdimen2\font plus
\BIBentryALTinterwordstretchfactor\fontdimen3\font minus
  \fontdimen4\font\relax}
\providecommand{\BIBforeignlanguage}[2]{{%
\expandafter\ifx\csname l@#1\endcsname\relax
\typeout{** WARNING: IEEEtran.bst: No hyphenation pattern has been}%
\typeout{** loaded for the language `#1'. Using the pattern for}%
\typeout{** the default language instead.}%
\else
\language=\csname l@#1\endcsname
\fi
#2}}
\providecommand{\BIBdecl}{\relax}
\BIBdecl

\bibitem{lucas2014}
A.~Lucas, ``Ising formulations of many {{NP}} problems,'' \emph{Frontiers in
  Physics}, vol.~2, Feb. 2014.

\bibitem{date2019}
P.~Date, R.~Patton, C.~Schuman, and T.~Potok, ``Efficiently embedding {{QUBO}}
  problems on adiabatic quantum computers,'' \emph{Quantum Information
  Processing}, vol.~18, no.~4, p. 117, Mar. 2019.

\bibitem{blekos2024}
K.~Blekos, D.~Brand, A.~Ceschini, C.-H. Chou, R.-H. Li, K.~Pandya, and
  A.~Summer, ``A review on {{Quantum Approximate Optimization Algorithm}} and
  its variants,'' \emph{Physics Reports}, vol. 1068, pp. 1--66, Jun. 2024.

\bibitem{punnen2022quadratic}
A.~P. Punnen, \emph{The quadratic unconstrained binary optimization
  problem}.\hskip 1em plus 0.5em minus 0.4em\relax Springer, 2022.

\bibitem{satoshi2008}
\BIBentryALTinterwordspacing
S.~Morita and H.~Nishimori, ``Mathematical foundation of quantum annealing,''
  \emph{Journal of Mathematical Physics}, vol.~49, no.~12, p. 125210, 12 2008.
  [Online]. Available: \url{https://doi.org/10.1063/1.2995837}
\BIBentrySTDinterwordspacing

\bibitem{pardalos1992}
P.~M. Pardalos and S.~Jha, ``Complexity of uniqueness and local search in
  quadratic 0\textendash 1 programming,'' \emph{Operations Research Letters},
  vol.~11, no.~2, p. 119, 1992.

\bibitem{gerlach2024fpga2}
T.~Gerlach, S.~Knipp, D.~Biesner, S.~Emmanouilidis, K.~Hauber, and
  N.~Piatkowski, ``Quantum optimization for fpga-placement,'' in
  \emph{Proceedings of the 2024 IEEE International Conference on Quantum
  Computing and Engineering (QCE)}.\hskip 1em plus 0.5em minus 0.4em\relax
  IEEE, 2024, pp. 637--647.

\bibitem{aguilera2024}
E.~Aguilera, J.~{de Jong}, F.~Phillipson, S.~Taamallah, and M.~Vos,
  ``Multi-{{Objective Portfolio Optimization Using}} a {{Quantum Annealer}},''
  \emph{Mathematics}, vol.~12, no.~9, p. 1291, Jan. 2024.

\bibitem{sakuler2025}
W.~Sakuler, J.~M. Oberreuter, R.~Aiolfi, L.~Asproni, B.~Roman, and J.~Schiefer,
  ``A real-world test of portfolio optimization with quantum annealing,''
  \emph{Quantum Machine Intelligence}, vol.~7, no.~1, p.~43, Mar. 2025.

\bibitem{lu2024}
Y.-C. Lu, C.-M. Fu, L.-P. Yu, Y.-J. Chang, and C.-R. Chang,
  ``Quantum-{{Inspired Portfolio Optimization In The QUBO Framework}},''
  \emph{arXiv:2410.05932 [q-fin]}, Nov. 2024.

\bibitem{stollenwerk2019flight}
T.~Stollenwerk, E.~Lobe, and M.~Jung, ``Flight gate assignment with a quantum
  annealer,'' in \emph{International Workshop on Quantum Technology and
  Optimization Problems}.\hskip 1em plus 0.5em minus 0.4em\relax Springer,
  2019, p.~99.

\bibitem{bauckhage2019}
C.~Bauckhage, N.~Piatkowski, R.~Sifa, D.~Hecker, and S.~Wrobel, ``A qubo
  formulation of the $k$-medoids problem,'' in \emph{Proceedings of the
  Conference on “Lernen, Wissen, Daten, Analysen” (LWDA)}.\hskip 1em plus
  0.5em minus 0.4em\relax CEUR-WS.org, 2019, p.~54.

\bibitem{mucke2023feature}
S.~M{\"u}cke, R.~Heese, S.~M{\"u}ller, M.~Wolter, and N.~Piatkowski, ``Feature
  selection on quantum computers,'' \emph{Quantum Machine Intelligence},
  vol.~5, no.~1, p.~11, 2023.

\bibitem{mucke2025}
\BIBentryALTinterwordspacing
S.~M{\"u}cke and T.~Gerlach, ``qubolite,'' 2023. [Online]. Available:
  \url{https://github.com/smuecke/qubolite}
\BIBentrySTDinterwordspacing

\bibitem{branke2008}
J.~Branke, K.~Deb, K.~Miettinen, and R.~S{\l}owi{\'n}ski, Eds.,
  \emph{Multiobjective {{Optimization}}: {{Interactive}} and {{Evolutionary
  Approaches}}}, ser. Lecture {{Notes}} in {{Computer Science}}.\hskip 1em plus
  0.5em minus 0.4em\relax Berlin, Heidelberg: Springer, 2008, vol. 5252.

\bibitem{miettinen1999nonlinear}
K.~Miettinen, \emph{Nonlinear multiobjective optimization}.\hskip 1em plus
  0.5em minus 0.4em\relax Springer Science \& Business Media, 1999.

\bibitem{xu2022}
H.~Xu, H.~{Ushijima-Mwesigwa}, and I.~Ghosh, ``Scaling {{Vehicle Routing
  Problem Solvers}} with {{QUBO-based Specialized Hardware}},'' in \emph{2022
  {{IEEE}}/{{ACM}} 7th {{Symposium}} on {{Edge Computing}} ({{SEC}})}, Dec.
  2022, pp. 381--386.

\bibitem{bouchmal2025}
O.~Bouchmal, B.~Cimoli, R.~Stabile, J.~J. Vegas~Olmos, C.~{Hernandez-Chulde},
  R.~Martinez, R.~Casellas, and I.~Tafur~Monroy, ``Quantum computing approach
  for multi-objective routing and spectrum assignment optimization,''
  \emph{Journal of Optical Communications and Networking}, vol.~17, no.~6, pp.
  B15--B27, Jun. 2025.

\bibitem{lin2025}
Y.-X. Lin, C.-Y. Xu, and C.~Wang, ``Multi-{{Objective Routing Optimization
  Using Coherent Ising Machine}} in {{Wireless Multihop Networks}},''
  \emph{arXiv:2503.07924 [quant-ph]}, Mar. 2025.

\bibitem{gunantara2018review}
N.~Gunantara, ``A review of multi-objective optimization: Methods and its
  applications,'' \emph{Cogent Engineering}, vol.~5, no.~1, p. 1502242, 2018.

\bibitem{matsumori2022}
T.~Matsumori, M.~Taki, and T.~Kadowaki, ``Application of {{QUBO}} solver using
  black-box optimization to structural design for resonance avoidance,''
  \emph{Scientific Reports}, vol.~12, no.~1, p. 12143, Jul. 2022.

\bibitem{marcelino2021efficient}
C.~G. Marcelino, G.~M. Leite, C.~A. Delgado, L.~B. de~Oliveira, E.~F. Wanner,
  S.~Jim{\'e}nez-Fern{\'a}ndez, and S.~Salcedo-Sanz, ``An efficient
  multi-objective evolutionary approach for solving the operation of
  multi-reservoir system scheduling in hydro-power plants,'' \emph{Expert
  Systems with Applications}, vol. 185, p. 115638, 2021.

\bibitem{dachert2022}
K.~D{\"a}chert, R.~Grindel, E.~Leoff, J.~Mahnkopp, F.~Schirra, and J.~Wenzel,
  ``Multicriteria asset allocation in practice,'' \emph{OR Spectrum}, vol.~44,
  no.~2, pp. 349--373, Jun. 2022.

\bibitem{miettinen2008}
K.~Miettinen, ``Introduction to {{Multiobjective Optimization}}:
  {{Noninteractive Approaches}},'' in \emph{Multiobjective {{Optimization}}:
  {{Interactive}} and {{Evolutionary Approaches}}}, ser. Lecture {{Notes}} in
  {{Computer Science}}, J.~Branke, K.~Deb, K.~Miettinen, and
  R.~S{\l}owi{\'n}ski, Eds.\hskip 1em plus 0.5em minus 0.4em\relax Berlin,
  Heidelberg: Springer, 2008, vol. 5252.

\bibitem{gass1955}
S.~Gass and T.~Saaty, ``The computational algorithm for the parametric
  objective function,'' \emph{Naval Research Logistics Quarterly}, vol.~2, no.
  1-2, pp. 39--45, 1955.

\bibitem{zadeh1963}
L.~Zadeh, ``Optimality and non-scalar-valued performance criteria,'' \emph{IEEE
  Transactions on Automatic Control}, vol.~8, no.~1, pp. 59--60, Jan. 1963.

\bibitem{ayodele2023}
M.~Ayodele, R.~Allmendinger, M.~{L{\'o}pez-Ib{\'a}{\~n}ez}, A.~Liefooghe, and
  M.~Parizy, ``Applying {{Ising Machines}} to {{Multi-objective QUBOs}},'' in
  \emph{Proceedings of the {{Companion Conference}} on {{Genetic}} and
  {{Evolutionary Computation}}}, ser. {{GECCO}} '23 {{Companion}}.\hskip 1em
  plus 0.5em minus 0.4em\relax New York, NY, USA: Association for Computing
  Machinery, Jul. 2023, pp. 2166--2174.

\bibitem{ayodele2023a}
M.~Ayodele, R.~Allmendinger, M.~{L{\'o}pez-Ib{\'a}{\~n}ez}, and M.~Parizy, ``A
  {{Study}} of {{Scalarisation Techniques}} for {{Multi-objective QUBO
  Solving}},'' in \emph{Operations {{Research Proceedings}} 2022}, O.~Grothe,
  S.~Nickel, S.~Rebennack, and O.~Stein, Eds.\hskip 1em plus 0.5em minus
  0.4em\relax Cham: Springer International Publishing, 2023, pp. 393--399.

\bibitem{lin2024}
X.~Lin, X.~Zhang, Z.~Yang, F.~Liu, Z.~Wang, and Q.~Zhang, ``Smooth
  {{Tchebycheff Scalarization}} for {{Multi-Objective Optimization}},''
  \emph{arXiv:2402.19078 [cs]}, Jul. 2024.

\bibitem{yu1973}
P.~L. Yu, ``A {{Class}} of {{Solutions}} for {{Group Decision Problems}},''
  \emph{Management Science}, vol.~19, no.~8, pp. 936--946, 1973.

\bibitem{deb2002}
K.~Deb, A.~Pratap, S.~Agarwal, and T.~Meyarivan, ``A fast and elitist
  multiobjective genetic algorithm: {{NSGA-II}},'' \emph{IEEE Transactions on
  Evolutionary Computation}, vol.~6, no.~2, pp. 182--197, Apr. 2002.

\bibitem{he2021}
L.~He, H.~Ishibuchi, A.~Trivedi, H.~Wang, Y.~Nan, and D.~Srinivasan, ``A
  {{Survey}} of {{Normalization Methods}} in {{Multiobjective Evolutionary
  Algorithms}},'' \emph{IEEE Transactions on Evolutionary Computation},
  vol.~25, no.~6, pp. 1028--1048, Dec. 2021.

\bibitem{boros2008max}
E.~Boros, P.~L. Hammer, R.~Sun, and G.~Tavares, ``A max-flow approach to
  improved lower bounds for quadratic unconstrained binary optimization
  (qubo),'' \emph{Discrete Optimization}, vol.~5, no.~2, p. 501, 2008.

\bibitem{goemans1995improved}
M.~X. Goemans and D.~P. Williamson, ``Improved approximation algorithms for
  maximum cut and satisfiability problems using semidefinite programming,''
  \emph{Journal of the ACM (JACM)}, vol.~42, no.~6, p. 1115, 1995.

\bibitem{karp1972}
R.~M. Karp, ``Reducibility among {{Combinatorial Problems}},'' in
  \emph{Complexity of {{Computer Computations}}}, R.~E. Miller, J.~W.
  Thatcher, and J.~D. Bohlinger, Eds.\hskip 1em plus 0.5em minus 0.4em\relax
  Boston, MA: Springer US, 1972, pp. 85--103.

\bibitem{boros1991}
E.~Boros and P.~L. Hammer, ``The max-cut problem and quadratic 0--1
  optimization; polyhedral aspects, relaxations and bounds,'' \emph{Annals of
  Operations Research}, vol.~33, no.~3, pp. 151--180, Mar. 1991.

\bibitem{biesner2022}
D.~Biesner, T.~Gerlach, C.~Bauckhage, B.~Kliem, and R.~Sifa, ``Solving {{Subset
  Sum Problems}} using {{Quantum Inspired Optimization Algorithms}} with
  {{Applications}} in {{Auditing}} and {{Financial Data Analysis}},'' in
  \emph{2022 21st {{IEEE International Conference}} on {{Machine Learning}} and
  {{Applications}} ({{ICMLA}})}, Dec. 2022, pp. 903--908.

\bibitem{SciPyProceedings_11}
A.~A. Hagberg, D.~A. Schult, and P.~J. Swart, ``Exploring network structure,
  dynamics, and function using {{NetworkX}},'' in \emph{Proceedings of the 7th
  python in science conference}, G.~Varoquaux, T.~Vaught, and J.~Millman, Eds.,
  Pasadena, CA USA, 2008, pp. 11--15.

\bibitem{harris2020array}
\BIBentryALTinterwordspacing
C.~R. Harris, K.~J. Millman, S.~J. van~der Walt et al., ``Array programming
  with {NumPy},'' \emph{Nature}, vol. 585, no. 7825, pp. 357--362, Sep. 2020.
  [Online]. Available: \url{https://doi.org/10.1038/s41586-020-2649-2}
\BIBentrySTDinterwordspacing

\bibitem{MuckePM19}
S.~M{\"{u}}cke, N.~Piatkowski, and K.~Morik, ``Hardware acceleration of machine
  learning beyond linear algebra,'' in \emph{International Conference on
  Machine Learning and Knowledge Discovery in Databases - Workshops}, ser.
  Communications in Computer and Information Science, P.~Cellier and
  K.~Driessens, Eds., vol. 1167.\hskip 1em plus 0.5em minus 0.4em\relax
  Springer, 2019, pp. 342--347.

\bibitem{zitzler1998}
E.~Zitzler and L.~Thiele, ``Multiobjective optimization using evolutionary
  algorithms --- {{A}} comparative case study,'' in \emph{Parallel {{Problem
  Solving}} from {{Nature}} --- {{PPSN V}}}, A.~E. Eiben, T.~B{\"a}ck,
  M.~Schoenauer, and H.-P. Schwefel, Eds.\hskip 1em plus 0.5em minus
  0.4em\relax Berlin, Heidelberg: Springer, 1998, pp. 292--301.

\end{thebibliography}



\end{document}